\documentclass[11pt]{article}

\usepackage[letterpaper,margin=1in]{geometry}
\usepackage[T1]{fontenc}
\usepackage{lmodern}
\usepackage{microtype}
\usepackage{graphicx}
\usepackage{booktabs}
\usepackage{multirow}
\usepackage{array}
\usepackage{tabularx}
\usepackage{siunitx}
\usepackage{enumitem}
\usepackage{placeins}
\usepackage{algorithm}
\usepackage{algorithmic}
\usepackage{amsmath,amssymb}
\usepackage{xcolor}
\usepackage{caption}
\usepackage{authblk}
\usepackage[numbers,sort&compress]{natbib}
\usepackage[hidelinks]{hyperref}
\usepackage[nameinlink,noabbrev]{cleveref}

\graphicspath{{figures/}}
\title{CAi Copilot: Reducing Operational Workload in Molecular Design through Intent-Driven Agentic Workflows}

\author[1]{Zhu Wang}
\author[2]{Jiangyu Chen}
\author[1]{Yingjun Shang}
\author[1]{Yuhui Yao}
\author[2]{Laiao Lu}
\author[1,2]{Tianfan Fu}
\author[1]{Na Zou}
\affil[1]{Shanghai Artificial Intelligence Laboratory, Shanghai, China}
\affil[2]{Nanjing University, Nanjing, China}
\date{}

\begin{document}
\maketitle

\begin{abstract}
Early-stage molecular design is an iterative process, not just a task of generating molecules. Researchers turn broad goals into design strategies, refine candidates, assess many properties, and gather evidence before synthesis and tests. AI methods can generate molecules, optimize several goals, predict properties, dock compounds, and account for synthesis. Yet these functions are spread across specialized tools. Experts must still coordinate each step, judge interim results, and integrate evidence. The central challenge is thus to turn research intent into adaptive, traceable runs grounded in scientific tools. We cast this challenge as intent-to-evidence molecular design workflow execution and present CAi Copilot, an expert-oriented agent with three linked layers. The Research Interface Layer turns intent into an executable plan. The Agent Reasoning Layer uses interim results to guide each run. The Execution Substrate supplies molecular tools, metrics, reusable utilities, and backend services. Across 45 tasks, CAi achieves the strongest overall performance, with an outcome score of 84.59, exceeding the next-best result by 18.07 points. Additional benchmarks test how CAi coordinates generation, screening, and multi-criteria evaluation, while exposing limits in long-horizon execution. These results show that CAi turns broad molecular-design intent into transparent, traceable workflows that connect interim decisions to candidate-level evidence.
\end{abstract}

\begin{center}
\small\textbf{Code:} \url{https://github.com/datamllab/CAi_copilot}
\end{center}

\section{Introduction}

A molecular candidate rarely emerges from one generation step. Early-stage design instead involves repeated proposal, assessment, filtering, ranking, and refinement. Researchers must explore vast chemical space while balancing activity, selectivity, ease of synthesis, safety, and drug-like properties~\cite{schneider2005computer,reymond2010chemical}. These goals often begin as broad research intent rather than a fully specified task. A request may combine a scaffold or seed molecule with a target, property limits, ranking rules, and external files. Progress toward synthesis and biological tests thus requires more than candidate generation. Researchers also need an executable strategy and enough evidence to compare candidates.

AI models now support many steps in this process. Molecular generators explore scaffolds and optimize structures, property models estimate key attributes, and docking methods assess target fit~\cite{du2024machine,loeffler2024reinvent,trott2010autodock}. Molecular language models also link chemical structures to text instructions and domain knowledge~\cite{edwards2022translation,fang2023molinstructions,zhang2024chemllm,yu2024llasmol}. Yet each model usually addresses a bounded task with fixed inputs and outputs. Given a broad request, researchers must still select the right operations, prepare inputs, connect diverse tools, review results, and revise the workflow after a failed step or a weak candidate set. Strong individual models, therefore, do not solve the workflow-level problem. 
Current molecular design methods provide powerful capabilities for individual tasks, yet lack a systematic framework that translates high-level research intent into coordinated execution and traceable, candidate-level evidence to support experts.

LLM-based scientific agents offer a basis for such coordination. Existing systems connect language reasoning with scientific tools, databases, and code execution~\cite{gottweis2026coscientist,ghareeb2026robin,aygun2026era,bran2024chemcrow,huang2025biomni}. Drug-discovery agents also cover therapeutic reasoning, target analysis, and multi-stage design workflows~\cite{drugagent2024,gao2025txagent,averly2025liddia}. Yet early-stage molecular design poses a distinct execution problem. Design intent may contain incomplete or context-sensitive constraints. Candidate development spans many models, file formats, metrics, and compute backends. Interim results may also alter later steps. Final advice must rest on evidence that experts can inspect. Experts remain responsible for scientific screening and experimental choices. 
A capable molecular design agent should enable end-to-end translation from research intent to coordinated execution and candidate-level evidence, extending beyond isolated tool use and molecule generation.

We formulate this setting as an intent-to-evidence molecular design workflow execution and present CAi Copilot. CAi turns broad design intent into an executable trajectory with candidate molecules, assessment records, rankings, and execution history. Three linked layers support this process. The Research Interface Layer grounds the molecular context and maps intent to a workflow plan. The Agent Reasoning Layer guides each run and updates the plan from interim results. The Execution Substrate supplies molecular tools, metrics, reusable utilities, domain skills, and backend services. This separation makes intent grounding, workflow control, and scientific execution explicit. Researchers can inspect the workflow, while computational tools handle repeated generation, assessment, and evidence assembly.

The main contributions of this paper are summarized as follows:
\begin{itemize}[leftmargin=*]
    \item We formulate early-stage molecular design as an intent-to-evidence workflow execution problem. This view shifts the output from an isolated molecule set to a traceable workflow record with candidate-level evidence.
    \item We present CAi Copilot, a three-layer agent for molecular design. CAi links intent grounding, result-driven control, code execution, and diverse tools. The framework unifies molecular generation, validation, assessment, filtering, ranking, and evidence reports.
    \item We evaluate CAi on 45 design tasks, further benchmarks, and 15 target-specific cases. The study covers agent execution and molecular outcomes. The results show strong intent-to-evidence performance and identify long-horizon execution as a remaining limit.
\end{itemize}

\section{Problem Formulation}
\label{sec:problem}

We first distinguish molecular generation from molecular design workflow execution. Molecular generation maps a given condition directly to candidate structures. Workflow execution starts with open-ended research intent. Such execution must organize molecular operations, respond to interim results, and gather evidence for expert review. We formulate the task as
\begin{equation}
    q=(I,\mathcal{C},\Gamma,\mathcal{T}),
\end{equation}
where $I$ is the design intent, $\mathcal{C}$ is the molecular context, $\Gamma$ holds the design constraints and evidence needs, and $\mathcal{T}$ is the available molecular tool library.

Given $q$, an agent maintains a workflow plan $P_t$ and a history $h_t$ of prior action--observation pairs. At step $t$, the agent selects and runs a molecular operation, then updates the remaining plan:
\begin{equation}
    \begin{array}{rcl}
        a_t &\sim& \pi(\cdot\mid h_t,P_t),\\
        o_{t+1} &=& \mathrm{Exec}(a_t;\mathcal{T}),\\
        P_{t+1} &=& \mathrm{Update}(P_t,o_{t+1}).
    \end{array}
\end{equation}
This update lets invalid molecular outputs, failed operations, or missing inputs alter later steps. The resulting action--observation sequence defines the execution trajectory $\tau$.

The output joins the trajectory with candidate-level evidence and explicit limits:
\begin{equation}
    \mathcal{E}=\{(m_i,\mathbf{z}_i,r_i,\rho_i)\}_{i=1}^{N},
    \qquad Y=(\mathcal{E},\tau,\mathcal{L}).
\end{equation}
For candidate $m_i$, $\mathbf{z}_i$ stores computed evidence, $r_i$ gives the candidate rank, and $\rho_i$ links each reported result to the supporting execution record. The set $\mathcal{L}$ stores missing evidence and failed operations. Let $S_I$, $S_\tau$, and $S_E$ denote intent satisfaction, trajectory validity, and evidence grounding. 
A successful run must satisfy all three conditions. In our evaluation, these conditions concern intent understanding and decomposition, workflow execution and operation coverage, and output validity, evidence consistency, and objective satisfaction. These aspects provide graded evidence of overall performance, whereas strict success requires complete task-level output matching.
\section{Agent Architecture}
\label{sec:agent_framework}

CAi Copilot bridges broad design intent and executable molecular computation. A research request may omit tool choices and execution details, while molecular tests reveal candidate quality only during the workflow. A fixed sequence cannot respond to invalid structures, unmet criteria, missing inputs, or failed operations. CAi therefore separates intent grounding, workflow control, and molecular execution. At the same time, CAi links each decision to a tool operation and candidate-level evidence.

\begin{figure*}[!t]
    \centering
    \includegraphics[
        page=1,
        width=0.65\textwidth,
        trim={149pt 81pt 172pt 75pt},
        clip
    ]{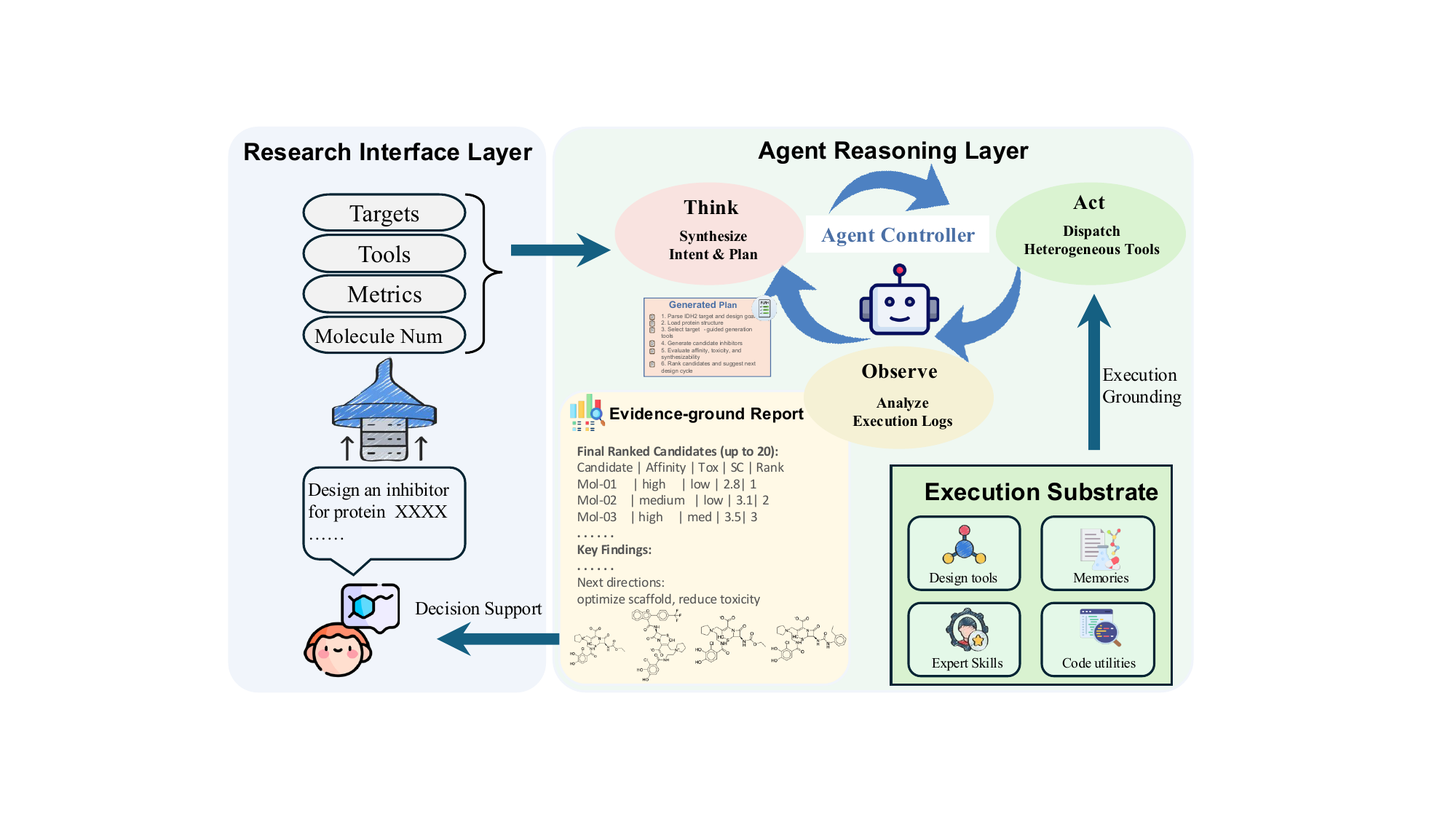}
    \caption{Overview of CAi Copilot. The three coordinated layers translate research intent into an executable plan, revise the workflow from molecular observations, and construct candidate-level evidence for expert review.}
    \label{fig:framework}
\end{figure*}

Figure~\ref{fig:framework} shows how the three layers coordinate the workflow. The Research Interface Layer builds a plan with explicit dependencies from the design request. The Agent Reasoning Layer and Execution Substrate then form a feedback loop that links molecular decisions, tool runs, and accumulated evidence. When the available tools permit, the reasoning layer may regenerate, refine, filter, or rerank candidates. Execution continues until the workflow obtains the requested evidence or records an unmet need. The three layers thus link workflow planning to iterative molecular design instead of acting as separate agents or a fixed pipeline.

\subsection{Research Interface Layer}
Given intent $I$, molecular context $\mathcal{C}$, and requirements $\Gamma$, the Research Interface Layer identifies the design goal, available inputs, required tests, ranking rules, and missing inputs. The layer then builds an initial plan
\begin{equation}
    P_0=\mathrm{Plan}(I,\mathcal{C},\Gamma)
       =(\mathcal{V}_0,\mathcal{D}_0),
\end{equation}
where $\mathcal{V}_0$ contains executable steps and $\mathcal{D}_0$ records the links between them. Each step has the form
\begin{equation}
    v_j=(g_j,u_j,x_j,p_j,e_j),
\end{equation}
where $g_j$ is the local goal, $u_j$ is the molecular operation, $x_j$ holds the required inputs, $p_j$ gives the input needs, and $e_j$ defines the expected evidence or completion signal.

The plan records links between steps, input needs, and expected evidence rather than only a tool list. A target-aware request may call for input checks, candidate generation, molecular filtering, target tests, and ranking by several metrics. Input checks expose unavailable operations before execution, while expected evidence links each operation to the design goal. During execution, $u_j$ maps to a suitable tool contract, $p_j$ is checked against the input needs in that contract, and $e_j$ contributes to the unmet evidence needs. The agent can revise the initial plan because later molecular results may change step settings, links, or unfinished work.

\subsection{Agent Reasoning Layer}
The Agent Reasoning Layer controls the closed-loop trajectory. At step $t$, the next operation depends on the current plan, prior steps, and candidate evidence gathered so far:
\begin{equation}
    a_t\sim\pi(\cdot\mid h_t,P_t,\mathcal{E}_t),
    \qquad o_{t+1}=\mathrm{Exec}(a_t;\mathcal{T}).
\end{equation}
The resulting observation updates the evidence state and the remaining evidence gap:
\begin{equation}
    \begin{array}{rcl}
    \mathcal{E}_{t+1}
      &=&\mathcal{E}_t\oplus\mathrm{Extract}(o_{t+1},a_t),\\
    \Delta_{t+1}
      &=&\mathrm{Req}(\Gamma)\setminus\mathrm{Cov}(\mathcal{E}_{t+1}),\\
    P_{t+1}
      &=&\mathrm{Update}(P_t,o_{t+1},\Delta_{t+1}).
    \end{array}
\end{equation}
Here, $\mathcal{E}_t$ is the evidence gathered through step $t$. Execution starts from $\mathcal{E}_0=\emptyset$ unless $\mathcal{C}$ already provides evidence for prior candidates. The operator $\oplus$ merges new molecular artifacts, computed values, warnings, and failures into the candidate records. $\mathrm{Req}(\Gamma)$ and $\mathrm{Cov}(\mathcal{E}_{t+1})$ denote the required and available evidence. Their difference $\Delta_{t+1}$ is the remaining evidence gap. This evidence-based control links generation to repeated optimization and screening. Invalid or duplicate candidates may trigger cleaning or regeneration, whereas constraint violations may trigger filtering or refinement. Partial results remain available. Missing inputs can defer a test or prompt an explicit request. Execution thus dynamically responds to intermediate molecular outcomes rather than following a predetermined, fixed tool sequence.

\subsection{Execution Substrate}
The Execution Substrate grounds agent decisions in molecular computation. The reasoning layer selects an operation, and the substrate determines whether the operation can run and how to return the result. Each tool $T_k\in\mathcal{T}$ follows an execution contract
\begin{equation}
    \kappa_k=(\mathcal{X}_k,\mathcal{Y}_k,
    \mathrm{Pre}_k,\mathrm{Exec}_k),
\end{equation}
where $\mathcal{X}_k$ and $\mathcal{Y}_k$ define accepted inputs and outputs, $\mathrm{Pre}_k$ lists prerequisites, and $\mathrm{Exec}_k$ invokes the backend. The contract explicitly exposes input readiness and downstream tool failures to the reasoning layer.

The substrate provides tools for molecular generation and optimization, evaluation metrics, reusable utilities, and domain workflow skills. Isolated backends support diverse runtime settings, checkpoints, hardware, file formats, and long jobs. Each tool call returns a standard observation with the execution status, molecular artifacts, computed values, warnings, and errors. This boundary keeps reasoning flexible, molecular operations transparent and inspectable, and complete runs reproducible.

\subsection{Cross-Layer Evidence and Decision Support}
Evidence is built across all three layers rather than in a separate component. Each observation updates the candidate record $(m_i,\mathbf{z}_i,r_i,\rho_i)$, where $\rho_i$ links a reported result to the supporting action and observation. Candidate records expose tool-derived values for the metrics in $\Gamma$. Filtering and ranking apply the requested thresholds, metric directions, and ranking rules to assess how well each option meets the design intent. Failed operations and unavailable evidence remain in $\mathcal{L}$. The same records guide later filtering, refinement, and ranking, which links workflow changes to the final candidate set.

A completed trajectory should satisfy the design intent, cover the requested evidence, and retain a valid execution history. If the available context or tools cannot meet a need, the workflow reports that limit. The final output contains ranked candidates, computational evidence, execution provenance, and known limits for expert review. Researchers remain responsible for chemical feasibility, biological relevance, experimental value, and later validation.

\section{Empirical Evaluation}
\label{sec:evaluation}

We evaluate CAi from workflow and molecular perspectives. CAiMD compares CAi with five agents on 45 tasks under a shared backbone and intent-to-evidence rubric. LIDDiA and MolBench retain their native protocols, while a target-specific case study examines designed molecules. We ask:

\begin{enumerate}[leftmargin=*,label=\textbf{RQ\arabic*:},nosep]
    \item How reliably do agents convert molecular-design intent into valid execution and grounded outcomes?
    \item Which workflow stages and task types limit reliable molecular-design execution?
    \item Does CAi remain competitive under independent molecular-agent benchmarks and native metrics?
    \item What does molecule-level analysis reveal about CAi-designed candidates in a target-specific case study?
\end{enumerate}

\subsection{Experimental Setup}

\paragraph{Tasks and Datasets.}
CAiMD contains 45 manually curated tasks organized into three 15-task molecular contexts: mutant IDH2 scaffold decoration, JAK-family dual-target design involving JAK1 and JAK2, and PDK1 lead optimization. Task templates abstract generation, optimization, scaffold- and target-based design, evaluation, filtering, and ranking operations from published design studies. We instantiate these templates with receptor structures, binding sites, co-crystallized ligands, leads, and scaffolds manually collected from DiffDec examples and associated PDB entries~\cite{xie2024diffdec}, JAK2 macrocycle studies~\cite{diao2023macrocyclization,hu2025macrocyclic}, and PDK1 studies using REINVENT 4 and curriculum learning~\cite{loeffler2024reinvent,guo2022curriculum}. The cited studies provide molecular contexts and source assets rather than agent-execution traces. Each task records the supplied inputs, required operations, output count, metrics, thresholds, ranking directions, and expected handling of invalid inputs.

\paragraph{Baselines.}
Five systems represent distinct agent paradigms: ChemCrow~\cite{bran2024chemcrow} for chemistry-specific tool use, Biomni~\cite{huang2025biomni} for biomedical research, AgentD~\cite{ock2026agentd} for modular drug discovery, and Codex~\cite{openai2025codexsystem} and Hermes~\cite{nous2026hermesagent} for general-purpose tool interaction. The comparison tests domain specialization, biomedical breadth, and general agent capability in end-to-end molecular-design workflows.

\paragraph{Implementation Details.}
For CAiMD, all agents use the same DeepSeek V4 Pro-compatible endpoint, prompts, inputs, isolated workspaces, and 2,400s budget. External benchmark comparisons use Qwen3.7-Max as the shared backbone. Native control logic and tools remain unchanged, so the experiments compare complete systems rather than tool-matched LLMs. All outputs are normalized and rescored with the same rubric. Unavailable capabilities and external failures are recorded without retries and separated from reasoning errors.

\paragraph{Evaluation Protocols.}
CAiMD evaluates intent recovery and decomposition (IU, TD), operation coverage and precision (TA, TP), input readiness and valid outputs (IP, VO), report consistency and the absence of unsupported analysis (RC, HF), and objective satisfaction (OS). Rules score verifiable criteria, while a blinded LLM judge scores semantic criteria; full definitions appear in the appendix. External evaluations retain released inputs and native metrics. SMDD-Bench~\cite{han2026smddbench} uses all 25 2D Pharmacophore Identification and 60 Fragment Assembly instances plus 30 randomly sampled Lead Optimization instances. LIDDiA~\cite{averly2025liddia} and MolBench~\cite{zhang2026molclaw} cover 30 targets and all 190 examples, respectively. 
Results for SMDD-Bench, LIDDiA, and MolBench are reported separately on different metrics and tasks.

\subsection{End-to-End Workflow Reliability (RQ1)}

Table~\ref{tab:agent_comparison} compares the complete intent-to-evidence trajectory across agents. The nine metrics trace performance from request interpretation to molecular outcomes. Their unweighted mean summarizes the profile without replacing analysis at each stage.

\begin{table*}[t]
\centering
\small
\setlength{\tabcolsep}{3pt}
\renewcommand{\arraystretch}{1.10}
\begin{tabular*}{\textwidth}{@{\extracolsep{\fill}}l*{9}{c}@{\hspace{7pt}}c@{}}
\toprule
\multirow{2}{*}{\textbf{Agent}}
& \multicolumn{2}{c}{\textbf{Intent}}
& \multicolumn{2}{c}{\textbf{Workflow}}
& \multicolumn{2}{c}{\textbf{Execution}}
& \multicolumn{2}{c}{\textbf{Evidence}}
& \multicolumn{1}{c}{\textbf{Outcome}}
& \multirow{2}{*}{\textbf{Avg.}} \\
\cmidrule(lr){2-3}\cmidrule(lr){4-5}\cmidrule(lr){6-7}\cmidrule(lr){8-9}\cmidrule(lr){10-10}
& \textbf{IU} & \textbf{TD} & \textbf{TA} & \textbf{TP} & \textbf{IP} & \textbf{VO} & \textbf{RC} & \textbf{HF} & \textbf{OS} & \\
\midrule
ChemCrow~\cite{bran2024chemcrow} & 95.56 & 51.11 & 17.78 & 37.78 & 48.67 & 25.93 & 53.80 & 54.44 & 57.93 & 49.22 \\
Biomni~\cite{huang2025biomni} & 95.56 & 59.11 & 37.78 & 44.89 & 61.78 & 30.31 & 62.41 & 56.11 & 64.89 & 56.98 \\
AgentD~\cite{ock2026agentd} & 95.56 & 41.11 & 6.67 & 36.56 & 48.67 & \underline{50.00} & 26.67 & 79.56 & 57.26 & 49.12 \\
Codex~\cite{openai2025codexsystem} & 95.56 & 72.44 & \textbf{68.89} & \underline{55.56} & \underline{70.22} & 34.91 & \textbf{81.18} & \textbf{81.78} & \underline{66.52} & \underline{69.67} \\
Hermes~\cite{nous2026hermesagent} & 95.56 & \textbf{79.44} & 28.89 & 36.00 & 30.00 & 26.67 & 26.67 & 49.56 & 55.78 & 47.62 \\
\addlinespace[1pt]
\textbf{CAi Copilot} & 95.56 & \underline{74.22} & \underline{57.78} & \textbf{65.33} & \textbf{79.33} & \textbf{67.49} & \underline{74.18} & \underline{80.89} & \textbf{84.59} & \textbf{75.49} \\
\bottomrule
\end{tabular*}
\caption{End-to-end agent performance on the 45 CAiMD tasks (\%). Metrics are grouped by evaluation dimension, and Avg. denotes their unweighted mean. Best results are bold and second-best results are underlined.}
\label{tab:agent_comparison}
\end{table*}

\paragraph{Performance separates after intent recognition.}
All systems score 95.56\% on intent understanding, while their overall means range from 47.62\% to 75.49\%. Hermes attains the best task-decomposition score but the lowest overall mean, showing that intent recovery and high-level planning do not determine end-to-end reliability.

\paragraph{Valid molecular outputs distinguish reliable execution.}
CAi improves the valid-tool-output rate over Codex by 32.58 percentage points, from 34.91\% to 67.49\%. Objective satisfaction increases by 18.07 points, from 66.52\% to 84.59\%, despite lower trajectory coverage and report consistency. The three domain-oriented baselines also score below Codex and CAi overall. These gaps arise mainly from coordinating domain operations into usable molecular evidence rather than from domain focus alone.

\subsection{Robustness to Metric Aggregation}

Table~\ref{tab:cai_overall} provides a complementary aggregation that first combines correlated metrics into capability dimensions and then assigns equal weight to the available dimensions and task outcome. This prevents the nine-metric mean from repeatedly rewarding closely related diagnostics. Efficiency is excluded because matched manual-time measurements are not yet available.

\begin{table}[t]
\centering
\small
\setlength{\tabcolsep}{6pt}
\renewcommand{\arraystretch}{1.08}
\begin{tabular}{@{}lccc@{}}
\toprule
\textbf{Agent} & \textbf{Capability} & \textbf{Outcome} & \textbf{Overall} \\
\midrule
ChemCrow & 47.52 & 57.93 & 49.25 \\
Biomni & 55.37 & 64.89 & 56.95 \\
AgentD & 45.95 & 57.26 & 47.84 \\
Codex & \underline{64.52} & \underline{66.52} & \underline{64.85} \\
Hermes & 47.33 & 55.78 & 48.74 \\
\textbf{CAi Copilot} & \textbf{73.94} & \textbf{84.59} & \textbf{75.71} \\
\bottomrule
\end{tabular}
\caption{Capability-level aggregation on the 45 CAiMD tasks (\%). Capability combines correlated metrics within workflow dimensions, while Overall assigns equal weight to the available dimensions and task outcome. Best results are bold and second-best results are underlined.}
\label{tab:cai_overall}
\end{table}

The cohort-level result preserves the main-table ordering: CAi leads Codex by 10.86 points overall and by 18.07 points on task outcome. Agreement between the two aggregation schemes indicates that the lead achieved by CAi is not an artifact of assigning equal weight to all nine diagnostics.

\subsection{Workflow-Stage Capability Diagnosis (RQ2)}

The five dimensions in Table~\ref{tab:agent_comparison} describe stages shared by tool-assisted molecular-design workflows and do not depend on agent architecture. Applying the same capability-level rubric to every system reveals where a trajectory loses reliability without rewarding a particular module design.

\paragraph{Execution is the largest cross-system differentiator.}
Intent recovery is nearly saturated, and the strongest planning scores do not preserve the overall ranking. The clearest separation between CAi and Codex instead occurs during execution: CAi improves input preparation by 9.11 points and valid tool output by 32.58 points. This concentration indicates that the principal advantage of CAi lies in converting workflow decisions into usable molecular artifacts, not in uniformly stronger reasoning or reporting.

\paragraph{No stage is sufficient in isolation.}
Hermes combines the best task decomposition with weak downstream execution, while Codex combines the best operation coverage and evidence scores with fewer valid outputs and lower objective satisfaction. Among the observed systems, reliable outcomes occur only when planning precision, executable inputs, valid outputs, and grounded reporting remain jointly strong. This pattern supports coordination across stages as the relevant capability, although aggregate scores do not establish a causal contribution for any single dimension.

\paragraph{CAi remains limited by coverage and evidence consistency.}
CAi does not lead in trajectory accuracy, report--tool consistency, or hallucination-free analysis. CAi executes selected operations reliably but may still omit required steps or give less consistent explanations than the strongest general agent. The task-family analysis tests whether these weaknesses grow in workflows with longer dependencies and iterative control.

Strict success falls most sharply on integrated workflows, whereas objective satisfaction remains stable. Long trajectories often retain grounded partial evidence but fail to complete every dependency, as detailed in the appendix.

\subsection{Generalization to External Benchmarks (RQ3)}

\subsubsection{SMDD.}


\begin{table}[t]
\centering
\small
\renewcommand{\arraystretch}{1.08}
\begin{tabular}{@{}l@{\hspace{14pt}}c@{\hspace{12pt}}c@{}}
\toprule
\textbf{Task} & \textbf{SMDD} & \textbf{CAi} \\
\midrule
2D Pharmacophore ID & 20.0 & \textbf{24.0} \\
Lead Optimization   & 50.0 & \textbf{63.3} \\
Fragment Assembly   & 0.0  & \textbf{1.7} \\
\bottomrule
\end{tabular}
\caption{Success rates of the SMDD Harness and CAi on the three
SMDD-Bench task types included in our evaluation (\%). Best results
are shown in bold.}
\label{tab:smdd_comparison}
\end{table}

We select three SMDD-Bench tasks that fall within the capability scope of the CAi agent and compare CAi with the SMDD Harness. We evaluate both systems on the complete 2D Pharmacophore Identification and Fragment Assembly test sets, comprising 25 and 60 instances, respectively, and on a randomly sampled subset of 30 Lead Optimization instances. Both systems use Qwen3.7-Max as the shared backbone while retaining their native control logic and tools.

Table~\ref{tab:smdd_comparison} demonstrates how CAi transfers across different molecular-design capabilities. On 2D Pharmacophore Identification, CAi improves the success rate from 20.0\% to 24.0\%, suggesting stronger molecular-pattern inference and screening execution. A larger gain appears on Lead Optimization, where the success rate increases from 50.0\% to 63.3\%, reflecting CAi's ability to coordinate iterative molecular modification, property evaluation, and multi-constraint filtering. On Fragment Assembly, CAi also solves instances that the SMDD Harness fails to complete, further extending its capability to tasks involving fragment composition and structural constraints. 

\subsubsection{LIDDiA.}

\label{sec:eval_liddia}

We compare both agents on 30 targets under a shared backbone, molecular budget, evaluator, and round-selection rule. The complete protocol is provided in the appendix.

Table~\ref{tab:liddia_cai} shows that CAi returns 78.5\% valid molecules and fills 39.2 of 50 slots on average. LIDDiA retains 14.4 valid molecules, or 28.7\% of the budget. CAi also leads every constraint-level pass rate, including a 28.7\% high-quality rate compared with 22.1\% for LIDDiA. Average quality is more balanced. LIDDiA leads NVT, QED, LRF, and macro-averaged Vina, while CAi achieves better synthetic accessibility, micro-averaged Vina, and macro diversity. Under the shared evaluator, the CAi advantage is therefore broad candidate coverage and constraint satisfaction rather than uniform dominance on every molecular-quality metric.

\begin{table}[t]
\centering
\small
\setlength{\tabcolsep}{0.5pt}
\renewcommand{\arraystretch}{1.05}

\begin{tabular*}{\columnwidth}{@{\extracolsep{\fill}}lcc@{\hspace{4pt}}cc@{}}
\toprule

\multirow{2}{*}{Metric}
& \multicolumn{2}{c}{Pass}
& \multicolumn{2}{c}{Quality}
\\

\cmidrule(lr){2-3}
\cmidrule(lr){4-5}

& \textbf{LIDDiA} & \textbf{CAi}
& \textbf{LIDDiA} & \textbf{CAi}
\\

\midrule

Generated
& -- / 50.0
& -- / 50.0
& --
& --
\\

Valid
& 28.7 / 14.4
& \textbf{78.5 / 39.2}
& --
& --
\\

\midrule

NVT $\uparrow$
& 25.9 / 12.9
& \textbf{51.9 / 25.9}
& \textbf{0.846 / 0.848}
& 0.823 / 0.812
\\

QED $\uparrow$
& 26.6 / 13.3
& \textbf{74.4 / 37.2}
& \textbf{0.737 / 0.733}
& 0.734 / 0.730
\\

LRF $\uparrow$
& 28.7 / 14.3
& \textbf{74.9 / 37.4}
& \textbf{3.995 / 3.991}
& 3.914 / 3.907
\\

SAS $\downarrow$
& 27.5 / 13.8
& \textbf{76.5 / 38.3}
& 2.488 / 2.546
& \textbf{2.275 / 2.302}
\\

Vina $\downarrow$
& 23.9 / 12.0
& \textbf{48.6 / 24.3}
& \textbf{-7.139} / -6.845
& -6.676 / \textbf{-6.860}
\\

HQ
& 22.1 / 11.1
& \textbf{28.7 / 14.3}
& --
& --
\\

DVS $\uparrow$
& --
& --
& 0.781 / \textbf{0.871}
& \textbf{0.794} / 0.808
\\

\bottomrule
\end{tabular*}
\caption{Budget-matched comparison with LIDDiA on 30 targets. Pass values report percentages and mean molecule counts, while quality values report macro and micro averages.}
\label{tab:liddia_cai}
\end{table}

\subsubsection{MolBench.}

\label{sec:eval_molclaw}

We compare CAi and MolClaw on all 190 MolBench examples using Qwen3.7-Max and the released evaluators. The execution setting is detailed in the appendix.

Table~\ref{tab:cai_molclaw_molbench} shows strong transfer on tool-verifiable tasks. CAi raises MS-1 accuracy from 0.1800 to 0.9600, reaches an editing accuracy of 0.9744 compared with 0.8718 for MolClaw, and matches both MS-2 measures. MolClaw remains stronger on receptor-dependent virtual screening, attaining 0.5200 Hit@3 while CAi reaches 0.4000. Limited visibility between local and remote structures constrained the docking workflow in this run. For optimization, CAi achieves a larger gain of 0.2263 and a higher success rate of 0.7949, although MolClaw preserves scaffolds more consistently. The native benchmark therefore confirms strengths of CAi in executable screening, molecular editing, and property improvement, while exposing limitations in structure-dependent tool coordination and scaffold-constrained optimization.

\begin{table}[t]
\centering
\small
\setlength{\tabcolsep}{2.5pt}
\renewcommand{\arraystretch}{0.88}

\begin{tabular}{@{}llrcc@{}}
\toprule
Bench. & Subtask & Metric & MolClaw & CAi \\
\midrule

\multirow{6}{*}{MS-1}
& \multirow{6}{*}{Screening}
& Accuracy $\uparrow$
& 0.1800 & \textbf{0.9600} \\

& & Precision $\uparrow$
& 0.3202 & \textbf{0.9800} \\

& & Recall $\uparrow$
& 0.4450 & \textbf{0.9700} \\

& & F1 $\uparrow$
& 0.3381 & \textbf{0.9733} \\

& & Specificity $\uparrow$
& 0.4400 & \textbf{0.9800} \\

& & Validity $\uparrow$
& 0.7800 & \textbf{1.0000} \\

\midrule

\multirow{2}{*}{MS-2}
& \multirow{2}{*}{Selection}
& Accuracy $\uparrow$
& \textbf{0.5405} & \textbf{0.5405} \\

& & Valid choice $\uparrow$
& \textbf{1.0000} & \textbf{1.0000} \\

\midrule

\multirow{5}{*}{MS-3}
& \multirow{5}{*}{Virtual screen.}
& Hit@3 $\uparrow$
& \textbf{0.5200} & 0.4000 \\

& & Avg. hits@3 $\uparrow$
& \textbf{0.7600} & 0.5600 \\

& & Top-3 hit rate $\uparrow$
& \textbf{0.2533} & 0.1867 \\

& & Avg. GT rank $\downarrow$
& 56.2267 & \textbf{55.4933} \\

& & Hit count $\uparrow$
& \textbf{13/25} & 10/25 \\

\midrule

\multirow{5}{*}{MO-Edit}
& Add
& Correct $\uparrow$
& 0.7000 & \textbf{1.0000} \\

& Delete
& Correct $\uparrow$
& 0.8889 & \textbf{1.0000} \\

& Substitute
& Correct $\uparrow$
& \textbf{0.9500} & \textbf{0.9500} \\

& Overall
& Correct $\uparrow$
& 0.8718 & \textbf{0.9744} \\

& Overall
& Validity $\uparrow$
& \textbf{1.0000} & \textbf{1.0000} \\

\midrule

\multirow{5}{*}{MO-Opt}
& \multirow{5}{*}{Overall}
& $\Delta$ $\uparrow$
& 0.1352 & \textbf{0.2263} \\

& & Success $\uparrow$
& 0.4872 & \textbf{0.7949} \\

& & Valid SMILES $\uparrow$
& 0.9744 & \textbf{1.0000} \\

& & Scaffold hard $\uparrow$
& \textbf{0.7179} & 0.4615 \\

& & Scaffold soft $\uparrow$
& \textbf{0.8894} & 0.7650 \\

\bottomrule
\end{tabular}
\caption{Performance on the 190-example MolBench benchmark. Best results are bold.}
\label{tab:cai_molclaw_molbench}
\end{table}

\subsubsection{Cross-Benchmark Analysis.}

Across CAiMD, SMDD-Bench, LIDDiA, and MolBench, CAi is strongest when workflow decisions must yield valid molecular outputs. CAiMD shows gains in output validity and objective satisfaction. SMDD-Bench shows improvements in all three tested categories, led by lead optimization, although fragment assembly remains near zero. LIDDiA demonstrates broader candidate coverage and higher constraint pass rates, while MolBench favors CAi in tool-verifiable screening, editing, and optimization. Recurring limits include omitted steps in integrated workflows, dependence on structure transfer in receptor-based screening, and reduced scaffold retention under aggressive optimization. These results associate reliable molecular-design agents with robust executable state transfer, systematic output validation, and constraint-aware candidate selection, but do not isolate the causal contribution of individual capabilities.



\subsection{Target-Specific Molecular Analysis (RQ4)}
\label{sec:case_study}

\paragraph{Case design.}
We analyze a 15-task JAK-family dual-target case involving JAK1 and JAK2 under a fixed structural context, with five variants for each target-, scaffold-, and lead-based entry point. The task composition is provided in the appendix. The tasks vary workflow scope, dependencies, invalid inputs, and constraint interpretation. Predicted JAK1 affinity is the primary objective. JAK1-over-JAK2 selectivity, synthetic accessibility, toxicity, ligand efficiency, and physicochemical properties remain supporting evidence rather than equally weighted objectives.

\begin{figure*}[t]
    \centering
    \begin{minipage}[t]{0.30\textwidth}
        \centering
        \includegraphics[width=\linewidth]{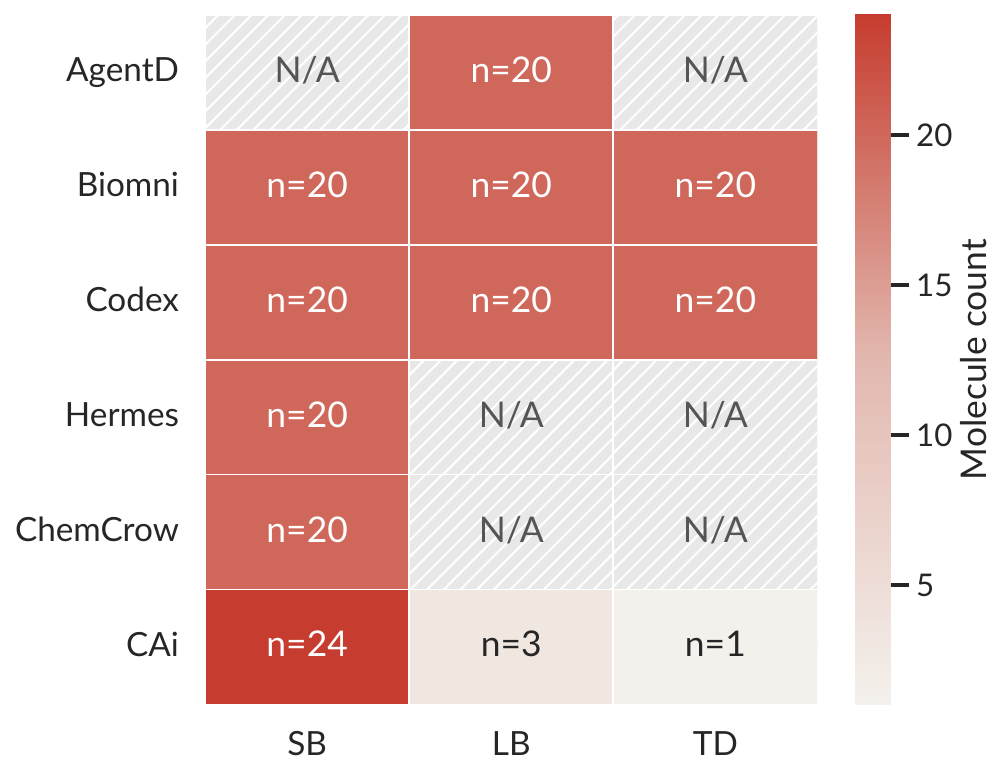}\\
        (a)
    \end{minipage}\hfill
    \begin{minipage}[t]{0.36\textwidth}
        \centering
        \includegraphics[width=\linewidth,trim=0 0 370 28,clip]{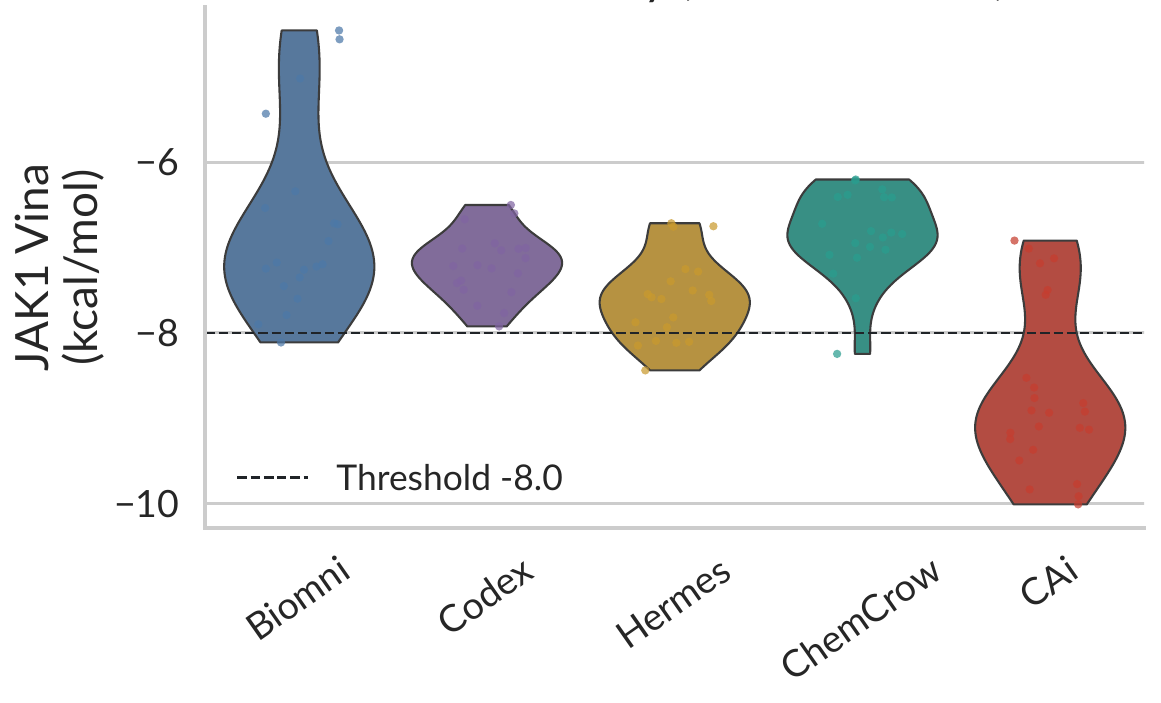}\\
        (b)
    \end{minipage}\hfill
    \begin{minipage}[t]{0.245\textwidth}
        \centering
        \includegraphics[width=\linewidth,trim=550 0 0 25,clip]{figures/case_study/sb_jak1_advantage.pdf}\\
        (c)
    \end{minipage}

    \begin{minipage}[t]{0.58\textwidth}
        \centering
        \includegraphics[width=\linewidth]{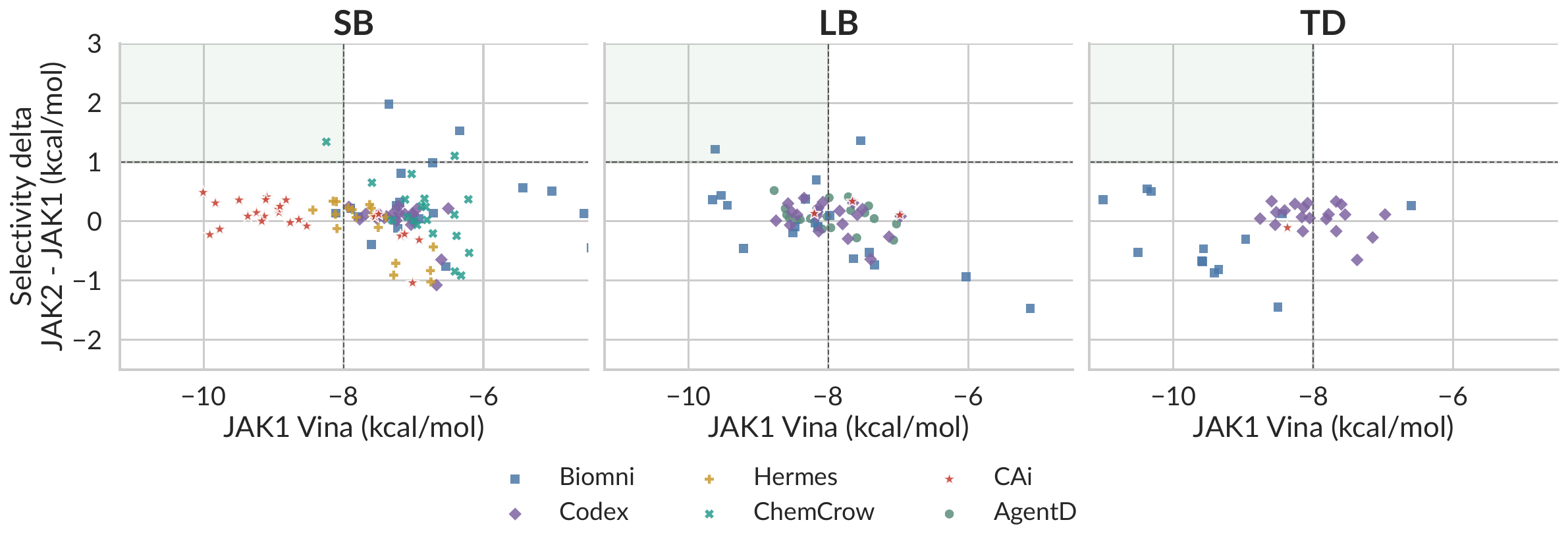}\\
        (d)
    \end{minipage}\hfill
    \begin{minipage}[t]{0.365\textwidth}
        \centering
        \includegraphics[width=\linewidth]{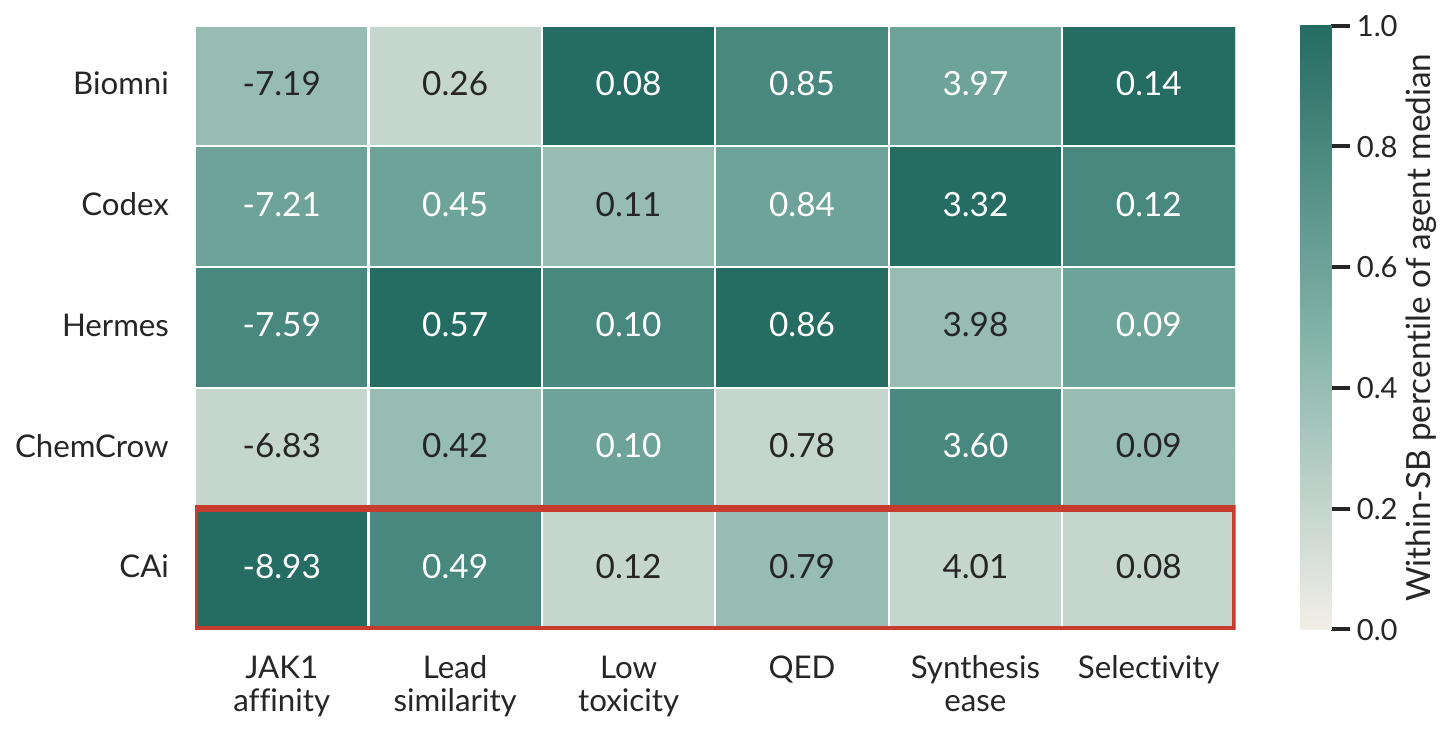}\\
        (e)
    \end{minipage}
    \caption{Workflow reachability and objective-conditioned molecular evidence in the JAK-family dual-target case involving JAK1 and JAK2. (a) Number of evaluable molecules returned for supported scaffold- (SB), lead- (LB), and target-based (TD) workflows. N/A denotes an unsupported combination. (b) Scaffold-based JAK1 Vina distributions. (c) Fraction satisfying Vina $\leq-8.0$ kcal/mol. (d) Affinity--selectivity distributions across the three entry points. (e) Median scaffold-based portfolio properties. The main molecular shift produced by CAi occurs on the requested JAK1 affinity objective, whereas reachability, JAK1-over-JAK2 selectivity, and auxiliary properties expose the remaining coverage and quality trade-offs. Results characterize predicted, selected portfolios and do not establish experimental activity or controlled generator superiority.}
    \label{fig:case_main}
\end{figure*}

\paragraph{Objective-aligned molecular enrichment.}
Figure~\ref{fig:case_main}a shows that CAi returns evaluable molecules for all three design entry points, although output counts vary across workflows. We therefore compare only supported outputs. In the matched scaffold-based setting, Figure~\ref{fig:case_main}b shows a shift in the predicted JAK1 affinity distribution toward stronger binding. CAi returns 24 selected candidates with a mean Vina score of $-8.706$ kcal/mol. As shown in Figure~\ref{fig:case_main}c, 18 candidates meet the prespecified $-8.0$ kcal/mol threshold, whereas only seven of 80 raw baseline molecules meet this threshold. Because CAi reports a selected portfolio whereas the baselines provide raw outputs, this comparison supports workflow-level prioritization rather than controlled generator superiority.

\paragraph{Portfolio trade-offs.}
Figure~\ref{fig:case_main}d shows that, across 201 molecules docked against both targets, CAi shifts most clearly along predicted JAK1 affinity, while JAK1-over-JAK2 selectivity remains concentrated near zero. Figure~\ref{fig:case_main}e further shows no uniform portfolio advantage in QED, toxicity, synthesis, similarity, or selectivity. Detailed distributions appear in the appendix. CAi therefore shows potential as an intent-conditioned prioritization agent. The workflow enriches candidates for the requested predicted-affinity objective while retaining evidence that makes molecular compromises visible to expert review.

\section{Related Work}

\noindent\textbf{Constraint-aware molecular generation.}
AI-based molecular design has progressed from unconstrained generation toward models that encode structural and chemical requirements. Pocket-conditioned methods model protein--ligand geometry to generate target-aware three-dimensional structures~\cite{feng2024lingo3dmol,peng2022pocket2mol,guan2024decompdiff}. Reaction-based generators instead build molecules from predefined transformations and available building blocks, improving synthesis awareness while retaining chemical diversity~\cite{cretu2024synflownet,koziarski2024rgfn}. These advances provide strong local capabilities under explicit inputs and objectives. However, a multi-objective design request can require several models and evaluation steps. Workflow construction must then connect compatible artifacts, combine affinity and molecular-property evidence, and respond to candidates that fail the stated criteria.

\noindent\textbf{Scientific workflow agents.}
Tool-augmented agents have expanded the scope of executable scientific workflows. ChemCrow connects language-model reasoning with chemistry tools, whereas Biomni combines retrieval, planning, and code execution across biomedical tasks~\cite{bran2024chemcrow,huang2025biomni}. Recent scientific agents extend this direction along complementary axes. Co-Scientist iteratively generates, critiques, and refines research hypotheses through a multi-agent architecture~\cite{gottweis2026coscientist}. Robin connects literature-grounded hypothesis generation with experimental data analysis in a lab-in-the-loop therapeutic workflow~\cite{ghareeb2026robin}. ERA uses language models and tree search to create and optimize empirical software for scorable scientific tasks~\cite{aygun2026era}. Drug-focused systems further coordinate in-silico candidate development: LIDDIA combines reasoning and evaluation, while MolClaw organizes molecular screening and optimization through hierarchical skills~\cite{averly2025liddia,zhang2026molclaw}. Collectively, these systems advance hypothesis generation, tool use, code execution, and iterative scientific feedback. A unified workflow for constructing diverse molecular design alternatives from open-ended expert intent remains underexplored. Such a workflow must compare candidates through multi-step molecular evaluations and preserve candidate-level evidence for expert selection.

\section{Conclusion}
\label{sec:conclusion}

Molecular design requires coordinated generation, assessment, filtering, and selection of candidates across multiple objectives and iterative stages.
We formulate this process as an intent-to-evidence workflow execution and present CAi Copilot. The Research Interface Layer, Agent Reasoning Layer, and Execution Substrate work together to plan, adapt, and run molecular workflows. Experiments show gains in valid molecular outputs and objective satisfaction. The target-specific analysis shows that CAi prioritizes affinity while preserving explicit evidence on selectivity and property trade-offs. CAi thus turns open-ended design intent into traceable candidate-level evidence for expert review. Experimental validation and final chemical judgment remain outside the computational workflow. Overall, CAi Copilot enables traceable, evidence-grounded molecular design under complex multi-objective property trade-offs, supporting experts with decision-making.

\FloatBarrier
\bibliographystyle{plainnat}
\bibliography{references}

\appendix
\section{CAiMD Construction and Data Provenance}
\label{app:data_construction}

\subsection{Source Context Extraction}

CAiMD is a workflow-execution evaluation suite rather than a collection of experimentally measured molecule labels. It contains three molecular-design contexts: mutant IDH2, paired JAK1/JAK2 design, and PDK1. Published studies define the design contexts and operation types, while the corresponding articles and PDB entries provide the receptor structures, binding sites, co-crystallized ligands, leads, and scaffolds used for task instantiation~\cite{xie2024diffdec,diao2023macrocyclization,hu2025macrocyclic,loeffler2024reinvent,guo2022curriculum}. The source studies do not provide agent-execution traces or reference plans.

For each context, receptor files and pocket parameters are recorded as structure-based inputs. Available SDF or SMILES records are parsed into molecular starting points. Lead-based cases retain a valid reference molecule, whereas scaffold-based cases use a core structure with explicit attachment points. Candidate attachment positions are treated as computational design inputs rather than experimentally validated synthetic routes. Table~\ref{tab:caimd_contexts} summarizes the retained resources and derived task inputs. The three contexts provide complementary starting conditions: PDK1 and IDH2 contribute multiple molecular records for lead- and scaffold-based instantiation, whereas the paired JAK1/JAK2 structures support target-aware affinity and selectivity workflows. This distinction ensures that CAiMD varies both molecular input form and structural context without treating the source molecules as ground-truth solutions.

\begin{table*}[t]
\centering
\scriptsize
\setlength{\tabcolsep}{4pt}
\renewcommand{\arraystretch}{1.10}
\begin{tabular}{p{0.11\linewidth} p{0.12\linewidth} p{0.25\linewidth} p{0.43\linewidth}}
\toprule
\textbf{Context} & \textbf{Structures} & \textbf{Molecular records} & \textbf{Derived task inputs} \\
\midrule
PDK1 & 2XCH & Five available active-molecule SDF records & Receptor and pocket configuration, lead SMILES, Murcko cores, and curated attachment-point scaffolds \\
IDH2 & 4JA8 & Six available ligand or active-molecule SDF records & Receptor and pocket configuration, lead SMILES, Murcko cores, and curated attachment-point scaffolds \\
JAK1/JAK2 & 6BBU \& 6BBV & Shared co-crystallized ligand D7D & Paired receptor and pocket configurations, a reference lead, and an attachment-point scaffold for affinity and selectivity workflows \\
\bottomrule
\end{tabular}
\caption{Scientific contexts retained during CAiMD construction. Molecular records provide starting points for task instantiation rather than ground-truth efficacy labels.}
\label{tab:caimd_contexts}
\end{table*}

\subsection{Factorized Task Instantiation}

The 45 cases follow a controlled $3\times3\times5$ design. Let $\mathcal{C}$ denote the three molecular contexts and $\mathcal{F}=\{\mathrm{TD},\mathrm{SB},\mathrm{LB}\}$ the target-, scaffold-, and lead-based design-entry families. For each context and entry family, five workflow configurations are instantiated:
\begin{equation}
    \mathcal{D}_{\mathrm{CAiMD}}
    =\bigcup_{c\in\mathcal{C}}\bigcup_{f\in\mathcal{F}}
    \{q_{c,f,v}:1\leq v\leq5\}.
\end{equation}
Thus, $|\mathcal{D}_{\mathrm{CAiMD}}|=3\cdot3\cdot5=45$. This construction varies workflow requirements while preserving the molecular context within each 15-task block. Table~\ref{tab:caimd_families} specifies how the same context is exposed through target-, scaffold-, and lead-based entry points. The families differ in their required prerequisites and validity checks, allowing the evaluation to test whether an agent adapts its workflow to the available scientific input rather than applying one fixed generation procedure.

\begin{table*}[t]
\centering
\scriptsize
\setlength{\tabcolsep}{4pt}
\renewcommand{\arraystretch}{1.10}
\begin{tabular}{p{0.12\linewidth} p{0.20\linewidth} p{0.31\linewidth} p{0.28\linewidth}}
\toprule
\textbf{Family} & \textbf{Initial context} & \textbf{Core workflow requirement} & \textbf{Input-specific checks} \\
\midrule
Target-based (TD) & Receptor structure and binding-site configuration & Generate candidates from structural context and, when requested, evaluate and filter them using target-related evidence & Receptor availability, pocket definition, docking configuration, and requested target comparisons \\
Scaffold-based (SB) & Core structure with explicit attachment points & Decorate or extend a constrained molecular core and apply the requested property or activity evaluations & Scaffold parseability, attachment-point validity, metric direction, and requested workflow scope \\
Lead-based (LB) & One or more reference molecules & Generate molecular analogs around a supplied lead and evaluate or prioritize them under the stated objectives & Molecular-format validity, lead availability, requested evaluation subset, and cross-step access to generated candidates \\
\bottomrule
\end{tabular}
\caption{Three design-entry families used to instantiate CAiMD. Each family contains five configurations in each molecular context.}
\label{tab:caimd_families}
\end{table*}

The five configurations within a context--family block are drawn from complementary workflow roles: (i) a complete multi-stage request, (ii) generation with evaluation explicitly excluded, (iii) evaluation or filtering of previously generated molecules, (iv) a missing or malformed prerequisite, and (v) a scope, threshold, metric-direction, or dependency stress condition. The molecular operation varies with the available context, but the evaluated behavior remains whether the agent follows the requested scope, respects dependencies, invokes compatible tools, and grounds the final report in returned evidence. TD/SB/LB therefore describe how a task enters the design process, whereas generation, evaluation, integrated-workflow, and error-handling labels describe the execution behavior required by the task. The two taxonomies are orthogonal.

\subsection{Reference Specifications and Quality Control}

The benchmark authors pair every prompt with a structured reference specification rather than a single target answer. The specification records (1) intent and requested output; (2) available receptor, molecule, scaffold, and file inputs; (3) required operations and step dependencies; (4) acceptable capability-level tool families; (5) metrics, directions, thresholds, and ranking rules; and (6) expected behavior when a prerequisite is absent or invalid. Capability-level references permit alternative valid implementations while preventing a final textual claim from substituting for missing execution evidence.

Case identifiers encode the molecular context, design-entry family, and workflow configuration. Consolidated records are checked for unique identifiers, resolvable inputs, valid molecular strings where validity is expected, explicit scaffold attachment points for scaffold generation, and consistent metric directions. Boundary cases intentionally violate one prerequisite and are retained only when the reference behavior is to stop, request the missing input, or report the limitation without fabricating molecules or scores. Generated candidates and predicted scores are evaluated as computational evidence, not as experimentally validated drug discoveries.

\FloatBarrier

\section{CAiMD Evaluation Protocol}
\subsection{Evaluation Metrics}
\label{app:metrics}

Let $R_i$ denote the reference set of required workflow operations for case $i$, and let $A_i$ denote the operations executed by an agent after mapping tool-specific calls to capability families. Following the terminology used in the main evaluation, we define rule-based trajectory accuracy and trajectory precision proxies as
\begin{equation}
    \mathrm{TA}_i=\frac{|R_i\cap A_i|}{|R_i|},
    \qquad
    \mathrm{TP}_i=\frac{|R_i\cap A_i|}{|A_i|}.
\end{equation}
These set-based proxies measure operation coverage and excess execution at the capability-family level; by themselves, they do not encode execution order or scientific validity. The final hybrid score uses a blinded LLM judge to assess dependency ordering, redundant or misplaced operations, and scientifically invalid actions that exact set matching cannot distinguish.

Let $U_i$ be the set of invoked tools whose required inputs are available and correctly formatted, and let $V_i\subseteq U_i$ be the invocations that return a usable molecular result. For all attempted invocations $C_i$, input-preparation success and valid-tool-output rate are
\begin{equation}
    \mathrm{IP}_i=\frac{|U_i|}{|C_i|},
    \qquad
    \mathrm{VO}_i=\frac{|V_i|}{|C_i|}.
\end{equation}
An invocation is usable only when the output can be parsed, linked to the requested candidates or task step, and consumed by a subsequent operation or final evidence report. Valid-tool-output rate is an end-to-end execution measure: because its denominator includes all attempted invocations, it captures both input-preparation failures and downstream tool failures rather than estimating conditional tool reliability.

For report grounding, let $G_i$ denote checkable claims in the final report, $C_i^{G}\subseteq G_i$ claims consistent with recorded tool outputs, and $H_i\subseteq G_i$ unsupported or contradicted analytical claims. We define
\begin{equation}
    \mathrm{RC}_i=\frac{|C_i^{G}|}{|G_i|},
    \qquad
    \mathrm{HF}_i=1-\frac{|H_i|}{|G_i|}.
\end{equation}
When no checkable claim is produced, the case is treated according to the expected-output specification for the task rather than assigned a perfect score.

Let $O_i$ denote the reference task objectives and $O_i^{+}$ the objectives satisfied by grounded outputs. Task objective satisfaction is
\begin{equation}
    \mathrm{OS}_i=\frac{|O_i^{+}|}{|O_i|}.
\end{equation}
Intent understanding is scored as exact agreement with the reference intent fields for rule evaluation. Task decomposition, semantic trajectory precision, partial input preparation, usable partial outputs, and hallucination-free interpretation are additionally assessed by a blinded judge on normalized $[0,1]$ rubrics. The final rubric and judge prompts will be released with the evaluation suite.

\subsection{Baseline Execution and Trace Normalization}

Each baseline is wrapped by an adapter that preserves the native control logic while writing a common artifact contract containing the final response, tool events, observations, execution status, and artifact paths. Codex and Hermes operate as general-purpose command-line agents. Biomni retains the biomedical tools and local worker dependencies provided by the native system. ChemCrow retains the chemistry-oriented LangChain tools, while AgentD enters the native target and lead extraction, generation, affinity, and ADMET pipeline. Unsupported capabilities are reported explicitly rather than emulated by a hidden replacement tool.

Before scoring, native trajectories are normalized into capability-level operations, tool-success indicators, generated-molecule evidence, reported metrics, and filtering or ranking decisions. A corrected AgentD adapter was rerun only because the initial harness rejected inputs before reaching the native pipeline. The corrected overlay retains downstream generation artifacts as well as genuine external-service failures. No task was rerun solely to improve a score. CAi is a previously completed reference run normalized and scored under the same evaluation rubric, rather than a sixth queue in the baseline supervisor.

\subsection{System-Level Comparison Protocol}
\label{app:system_boundaries}

CAiMD evaluates each baseline in its native executable configuration while preserving its released control logic and public tool implementation. To improve comparability, all systems receive matched task inputs and model budgets, and their trajectories are normalized under a shared artifact contract and scoring rubric. The resulting scores characterize practical end-to-end workflow reliability across complete systems, with capability availability and execution conditions treated as part of the evaluated system setting.

\FloatBarrier

\section{Target-Specific Case-Study Details}
\label{app:case_diagnostics}

This section provides molecular-property diagnostics for the target-specific case study. The supplementary analyses characterize factors that are not captured by absolute docking score alone.

\begin{figure*}[!t]
    \centering
    \captionsetup{skip=2pt}
    \includegraphics[width=0.98\textwidth,trim=8 8 8 86,clip]{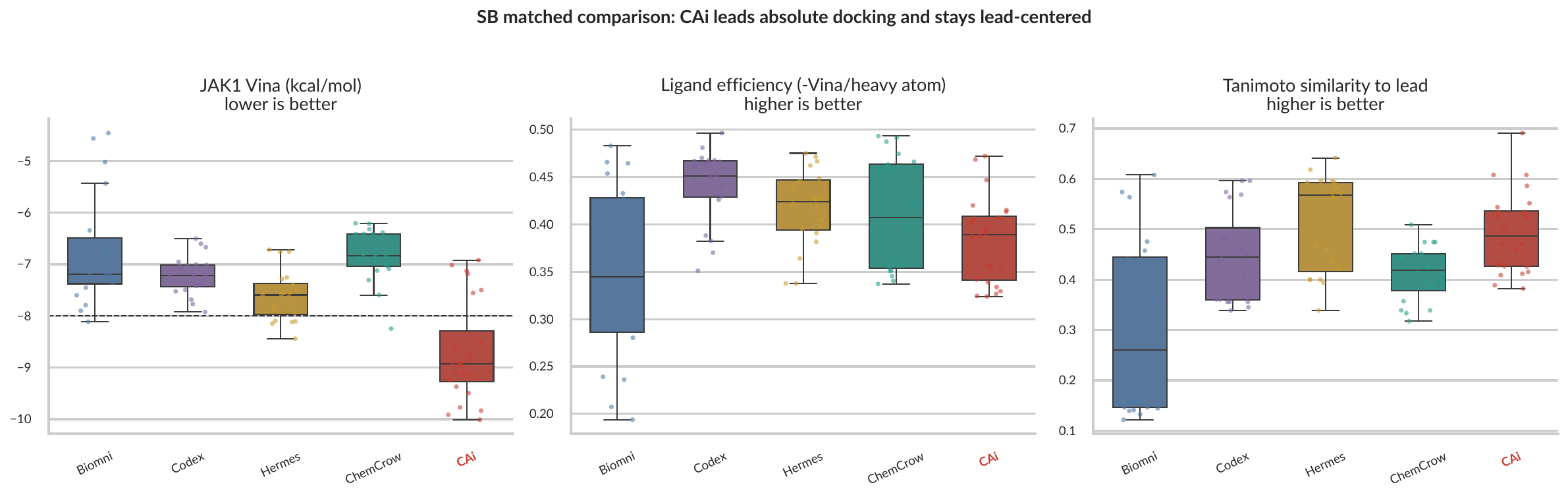}
    \caption{Scaffold-based distributions of predicted JAK1 Vina score, docking-based ligand efficiency, and Tanimoto similarity to the supplied lead. Under the affinity-prioritized case specification, CAi shows the lowest observed median JAK1 Vina score. Ligand efficiency and lead similarity provide complementary characterization of the selected portfolio.}
    \label{fig:case_efficiency}
\end{figure*}

\begin{figure*}[!t]
    \centering
    \captionsetup{skip=2pt}
    \includegraphics[width=0.96\textwidth,trim=8 0 8 44,clip]{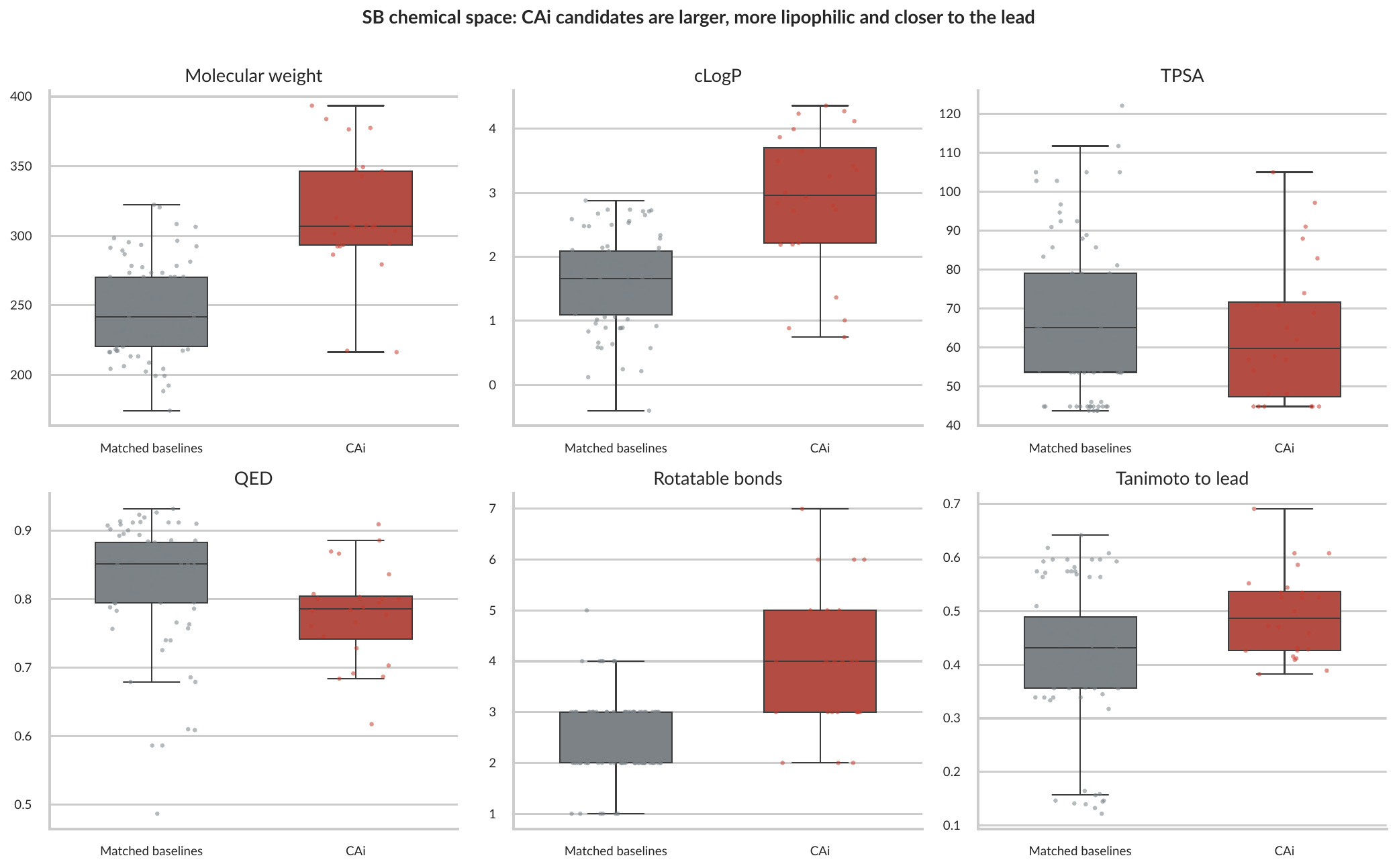}
    \caption{Physicochemical distributions for workflow-matched scaffold-based outputs. Molecular weight, cLogP, TPSA, QED, rotatable bonds, and lead similarity provide context for interpreting differences in docking-based affinity.}
    \label{fig:case_properties}
\end{figure*}

\subsection{Molecular Property Diagnostics}

Figure~\ref{fig:case_efficiency} compares the workflow-matched scaffold-based candidate sets using predicted JAK1 Vina score (lower is better), docking-based ligand efficiency computed as $-\mathrm{Vina}/N_{\mathrm{heavy}}$ (higher is better), and Tanimoto similarity to the supplied lead (higher is better). CAi produces the most favorable JAK1 docking-score distribution while retaining a lead-centered candidate set. Its ligand-efficiency distribution is not uniformly dominant, indicating that the docking improvement reflects objective-aligned prioritization rather than a universal gain across size-normalized criteria.

Figure~\ref{fig:case_properties} further characterizes the chemical space associated with this result. Relative to the matched baseline pool, CAi candidates tend to have higher molecular weight and cLogP, more rotatable bonds, and greater similarity to the supplied lead, while remaining within the observed QED and TPSA ranges. Together, the two figures show that CAi follows the affinity-prioritized case specification and exposes the accompanying molecular-property profile for expert review, rather than presenting the docking score as an isolated claim.

\section{External Benchmark Protocols}

\subsection{SMDD-Bench}
\label{app:smdd_protocol}
We evaluate the three SMDD-Bench task types that fall within CAi's current capability scope~\cite{han2026smddbench}. The comparison uses all 25 instances of 2D Pharmacophore Identification, all 60 instances of Fragment Assembly, and the same randomly sampled subset of 30 Lead Optimization instances for both systems. In 2D Pharmacophore Identification, each instance provides a target sequence together with 10 active and 10 inactive molecules, and requires an executable Python function that identifies the inferred pharmacophore; the released evaluator tests this function on hidden actives and inactives. Lead Optimization provides a reference protein--ligand complex, properties to optimize, properties to hold constant, and hard molecular constraints, and evaluates the submitted molecule with the released RDKit, ADMET-AI, and Boltz2 checks. Fragment Assembly provides one or two three-dimensional fragments positioned in a protein pocket and requires a single drug-like molecule that contains and, when necessary, links the fragments while satisfying the released structural and binding checks.

We compare CAi with the original SMDD Harness using Qwen3.7-Max as the common language-model backbone. Both systems receive the same task instances, input files, and allowed task information, and their final submissions are scored by the official task-specific evaluators. The SMDD Harness retains its released minimalist ReAct control scaffold and tool environment, whereas CAi retains its native workflow controller and molecular tools; the experiment is therefore a same-backbone, system-level comparison rather than a tool-matched ablation. For each task type, success rate is the fraction of instances whose final submission passes all evaluator-defined requirements. We do not assign partial credit or map the native SMDD-Bench outcomes to the CAiMD rubric.

\subsection{LIDDiA}
\label{app:liddia_protocol}
CAi and LIDDiA are evaluated on 30 targets with Qwen3.7-Max, identical task inputs and evaluator, at most three rounds, 50 candidates per round, and 50 evaluated slots per target. The budget is matched at the candidate level because a CAi round covers a complete proposal--evaluation cycle, whereas a LIDDiA action may perform generation, optimization, or screening. Molecules removed by final screening count as failed slots under the fixed denominator. The same constraint-based rule selects the best completed round for both agents. Pass results report the percentage and mean count under the 50-slot denominator. Molecular-quality results report macro and micro averages. Metric directions follow the released evaluation, and unavailable entries are not scored.

\subsection{MolBench}
\label{app:molbench_protocol}
CAi and the released MolClaw reference are evaluated on all 190 MolBench examples spanning molecular screening, selection, editing, and optimization. Both use Qwen3.7-Max and the original task inputs and evaluators. CAi runs the control process locally and invokes chemistry tools on a remote node. Native metrics are retained for each subtask and reported separately rather than combined across heterogeneous capabilities. Unavailable receptor structures on the remote node are recorded as execution limitations rather than replaced with alternative inputs.

\FloatBarrier

\end{document}